\documentclass[journal]{IEEEtran}

\ifCLASSINFOpdf
\else
   \usepackage[dvips]{graphicx}
\fi
\usepackage{url}
\usepackage{graphicx}
\usepackage{amssymb}
\usepackage{amsfonts}
\usepackage{xcolor}
\usepackage{balance}
\usepackage{ifthen}

\newcommand{\eg}{\textit{e.g.}, }
\newcommand{\ie}{\textit{i.e.}, }

\begin{document}

\title{Superpixel Segmentation using Dynamic and Iterative Spanning Forest}

\author{Felipe C. Bel{\'e}m, Silvio J. F. Guimar{\~a}es, and Alexandre X. Falc{\~a}o
    \thanks{Submitted in October 4th, 2019. The authors thank CNPq (303808/2018-7, 131000/2018-7) and FAPESP (2014/12236-1) for the financial support.}
    \thanks{F.C. Bel{\'e}m and A.X. Falc{\~a}o are with the Laboratory of Image Data Science, Institute of Computing, University of Campinas (UNICAMP), 
    (felipe.belem@ic.unicamp.br, afalcao@ic.unicamp.br).} 
    \thanks{S.J.F. Guimar{\~a}es is 
    at the Pontificial Catholic University of Minas Gerais (PUC-Minas), 
    (sjamil@ pucminas.br).
    }
}

\markboth{Journal of \LaTeX\ Class Files, Vol. 14, No. 8, August 2015}
{Shell \MakeLowercase{\textit{et al.}}: Bare Demo of IEEEtran.cls for IEEE Journals}
\maketitle

\begin{abstract}
As constituent parts of image objects, superpixels can improve several higher-level operations. However, image segmentation methods might have their accuracy seriously compromised for reduced numbers of superpixels. We have investigated a solution based on the \emph{Iterative Spanning Forest} (ISF) framework. In this letter, we present \emph{Dynamic} ISF (DISF) --- a method based on the following steps. (a) It starts from an image graph and a seed set with considerably more pixels than the desired number of superpixels. (b) The seeds compete among themselves, and each seed conquers its most closely connected pixels, resulting in an image partition (spanning forest) with connected superpixels. In step (c), DISF assigns \emph{relevance} values to seeds based on superpixel analysis and removes the most irrelevant ones. Steps (b) and (c) are repeated until the desired number of superpixels is reached. DISF has the chance to reconstruct relevant edges after each iteration, when compared to region merging algorithms. As compared to other seed-based superpixel methods, DISF is more likely to find relevant seeds. It also introduces dynamic arc-weight estimation in the ISF framework for more effective superpixel delineation, and we demonstrate all results on three datasets with distinct object properties.
\end{abstract}


\IEEEpeerreviewmaketitle

    \section{Introduction} \label{sec:intro}
        
        \begin{figure*}[ht!]
            \centering
            \includegraphics[width = 0.85\textwidth]{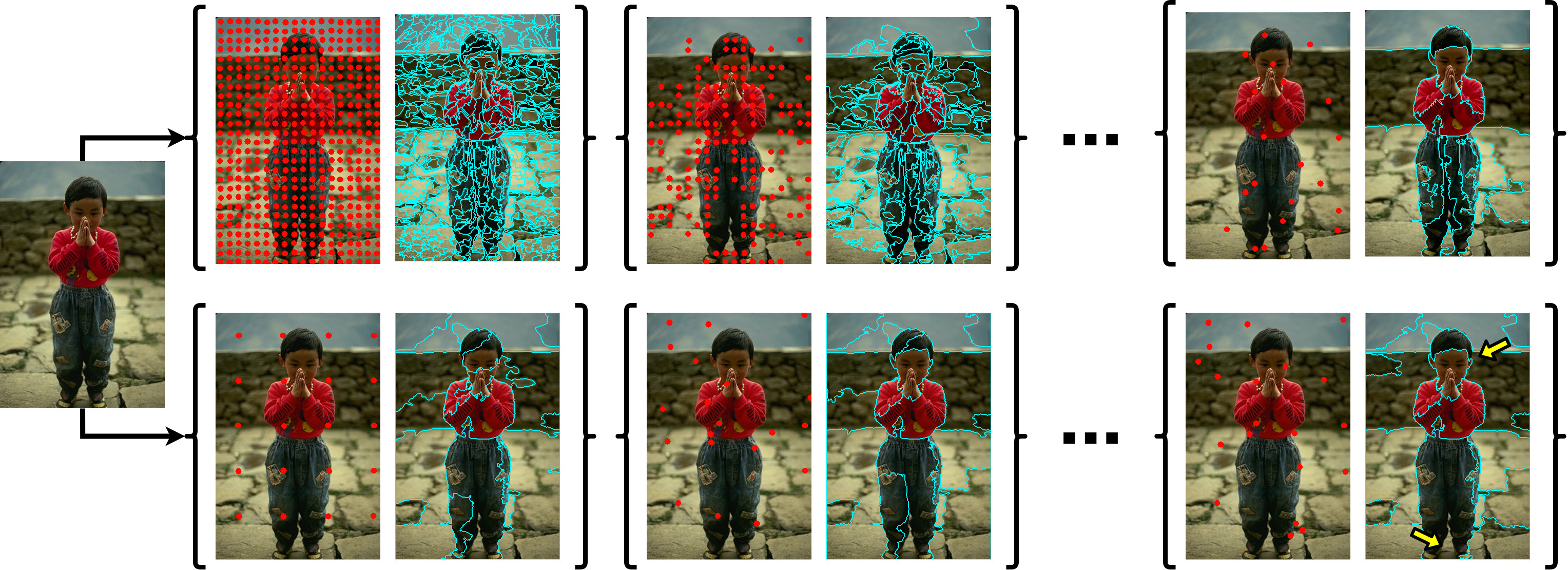}
            \caption{Segmentation results at each iteration to achieve 20 superpixels using DISF (top) and a competitive ISF-based method~\cite{vargas2018isf} (bottom). DISF starts from 200 seeds and ends with exactly 20 superpixels. Red dots indicate seed pixels, and yellow arrows indicate major segmentation errors. }
            \label{fig:two_pipelines} \vspace*{-0.2cm}
        \end{figure*}
        
        \IEEEPARstart{S}{uperpixels} are groups of connected pixels that share similar characteristics according to a predicate, being relevant in several applications: real-time image processing~\cite{morerio2014optimizing}, saliency detection~\cite{saliency2014tong}, and medical image analysis~\cite{li2018skin}.
        
        In \cite{stutz2018superpixels}, superpixel segmentation methods are classified as path-based~\cite{vargas2018isf} and clustering-based~\cite{achanta2012slic,liu2018intrinsic,li2015superpixel} approaches. A major group among recent methods adopts a three-step pipeline based on seed pixels: (i) initial seed sampling; (ii) superpixel delineation; and (iii) seed recomputation. In (i), the methods often aim a number of equidistant seeds equal to the desired number of superpixels. In (ii), superpixels are delineated based on some similarity criterion, which incorporates a pixel to the superpixel of its most similar seed. In (iii), the seeds may change location based on some homogeneity criterion, and steps (ii) and (iii) repeat to improve superpixel segmentation in a few iterations.
        
        One drawback of the above pipeline is limiting the initial seed set to the desired number of superpixels, which might considerably reduce the chances of finding seeds that will lead to accurate delineation of the important object edges --- the \textit{relevant seeds}. As a result, the methods might have their accuracy seriously compromised for reduced numbers of superpixels. We have investigated a solution based on the \emph{Iterative Spanning Forest} (ISF) framework~\cite{vargas2018isf}. In ISF, the image is a graph whose pixels are the nodes and arcs connect adjacent pixels. For the given seed set and path-cost function, the  \emph{Image Foresting Transform} (IFT) algorithm~\cite{falcao2004ift} computes a spanning forest in the graph such that each tree is a connected superpixel rooted at one seed. The previous ISF-based methods adopt the same three-step pipeline with different solutions for steps (i)-(iii), including different path-cost functions. However, they suffer from the same problem of seed set limited to the number of desired superpixels (see the bottom row of Figure~\ref{fig:two_pipelines}).
        
        In this letter, we introduce a new three-step pipeline for ISF-based methods, increasing the chances of finding relevant seeds, and then being more effective for reduced numbers of superpixels. We also propose one method, called \emph{Dynamic} ISF (DISF), that incorporates the concept of dynamic arc-weight estimation~\cite{bragantini2019dynift} into ISF-based superpixel segmentation. The idea is to extract information from each growing tree (superpixel) to estimate the path cost to each new pixel during the IFT algorithm. The new pipeline starts from (a) an image graph and a seed set with significantly higher size than the desired number of superpixels. In step (b), the IFT algorithm executes for superpixel delineation using dynamic arc-weight estimation in the path-cost function. In step (c), DISF assigns a relevance value to each seed based on superpixel analysis and removes the most irrelevant ones from the seed set. It then repeats steps (c) and (b) until it reaches the desired number of superpixels.
        
        The oversampling in step (a) increases the chances to include relevant seeds in the initial set and so capture all important object edges in step (b), when seeds compete among themselves and each seed conquers its most closely connected pixels (see the top row of Figure~\ref{fig:two_pipelines}). The dynamic arc-weight estimation in step (b) adapts the path-cost function to consider the mid-level image properties of each growing tree rather than the usual pixel properties for a more reliable estimation of the cost to incorporate a new pixel. The role of step (c) is to hold the most relevant seeds in the set, such that the main object edges can always be reconstructed in step (b). 

\begin{figure}[b!]
            \centering
            \includegraphics[width = 0.27\textwidth]{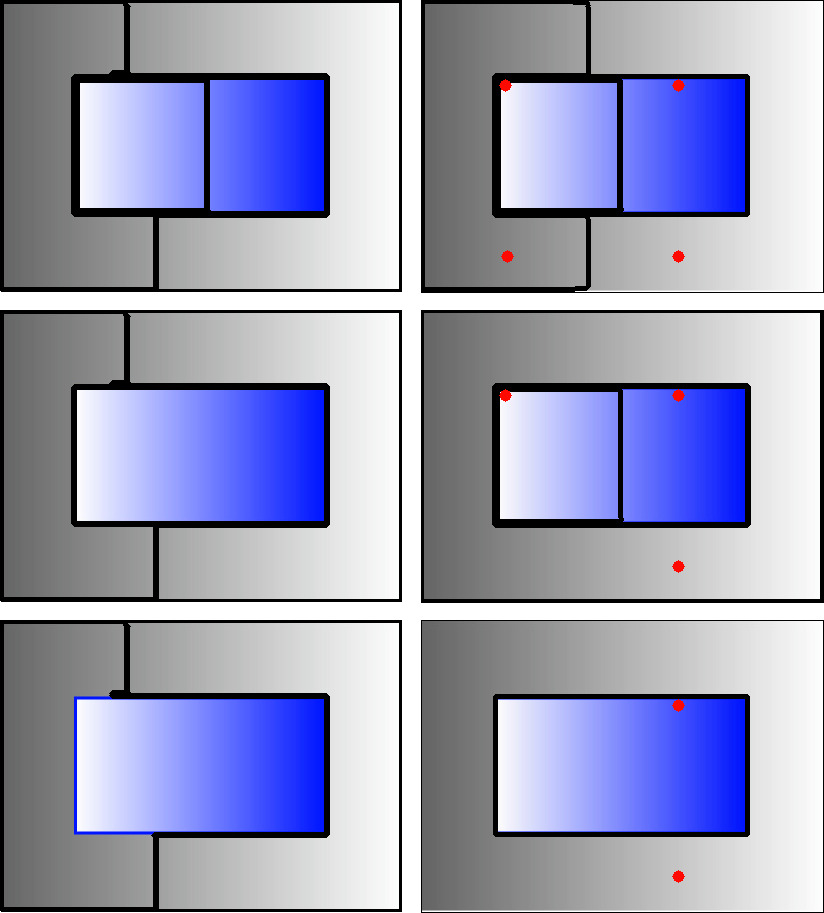}
            \caption{Segmentation results by region merging~\cite{wei2018sh} (left) and DISF (right) along iterations that result 4, 3, and 2 superpixels from top to bottom, respectively. The seeds are indicated by red dots in DISF.} \vspace*{-0.2cm}
            \label{fig:diff_hierar}
        \end{figure}

        It is important to notice that our approach cannot be classified as a hierarchical superpixel segmentation method, since it does not generate a hierarchy of segmentations, neither it respects the causality and locality principles stated in ~\cite{guigues2006scale}. In contrast to hierarchical region-merging algorithms --- which propagate delineation and merging errors to upper levels in the hierarchy ---, DISF can reconstruct object edges by promoting the competition amongst relevant seeds at every iteration (Figure~\ref{fig:diff_hierar}).
        
        In summary, our contributions are: (1) a new three-step pipeline for seed-based superpixel segmentation, which aims to include relevant seeds in the initial seed set and retain them during the process; (2) a new ISF-based method, named DISF, which relies on the new pipeline; (3) rules to estimate seed relevance and number of irrelevant seeds for removal at each iteration, such that the desired number of superpixels is always achieved at the end of the process; and (4) the incorporation of dynamic arc-weight estimation in IFT-based superpixel delineation for more effective segmentation. Some of these contributions can also benefit ISF-based methods~\cite{belem2019oisf, galvao2018risf,martins2019symmisf,castelo-fernandezLNCS19} recently developed for distinct applications.
        
        Section~\ref{sec:disf} presents DISF, and the experimental results on three image datasets with different object properties are shown in Section~\ref{sec:results}. Section~\ref{sec:concl} states conclusion and provides directions to future work.
        
    \section{Dynamic and Iterative Spanning Forest (DISF)} \label{sec:disf}
        We present here the three steps of DISF: (a) seed oversampling, (b) IFT-based superpixel delineation, and (c) seed set reduction.
        
        \newcommand{\domain}{D_I}
        \newcommand{\intensity}{\textbf{I}}
        \newcommand{\adjac}[1]{\mathcal{#1}}
        \newcommand{\seedset}{\mathcal{S}}
        \newcommand{\nodeset}{\mathcal{N}}
        \newcommand{\treesset}{\mathcal{T}}
        An image is a pair $(\domain, \intensity)$, being $\domain$ the set of pixels and $\intensity(p)$ a mapping that assigns local image attributes to each $p \in \domain$. We use the Lab color space, but those attributes could also be derived from image filtering. For the given node set $\nodeset \subseteq \domain$ and \textit{adjacency relation} $\adjac{A} \subset \nodeset \times \nodeset$, one can define an \textit{image graph} $G$ as $(\nodeset,\adjac{A},\intensity)$. In this work, we consider 2D images and the 8-neighborhood adjacency relation.
        
        \subsection{Seed Oversampling} \label{subsec:seedsampl}
            \begin{figure*}[t!]
                \centering
                \begin{tabular}{c c c c c}
                    \includegraphics[width=0.18\textwidth, trim =0 60 0 50, clip]{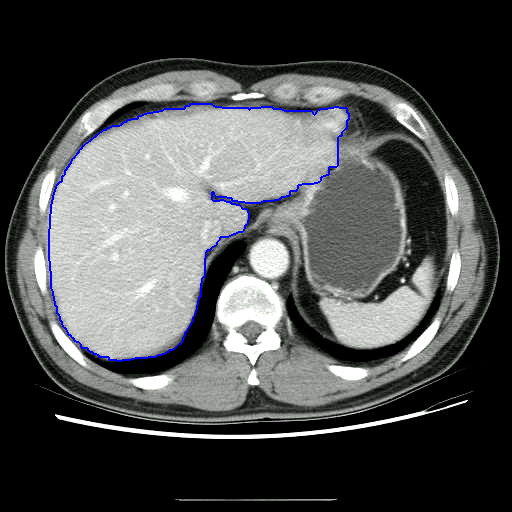} & \includegraphics[width=0.15\textwidth]{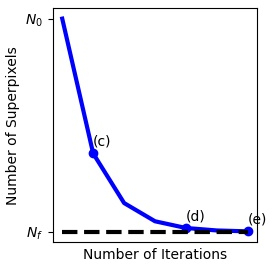} &
                     \includegraphics[width=0.18\textwidth, trim=0 60 0 50, clip]{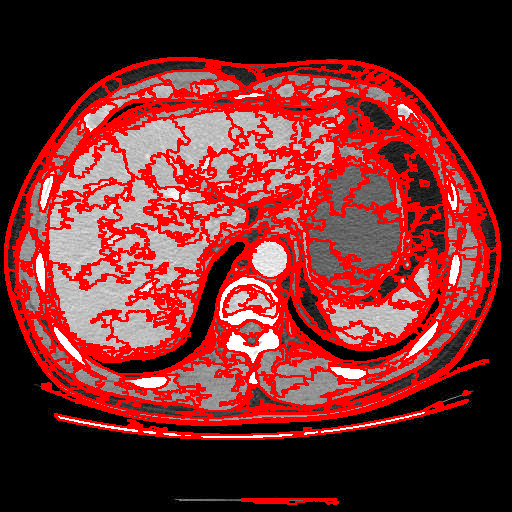} &
                     \includegraphics[width=0.18\textwidth, trim=0 60 0 50, clip]{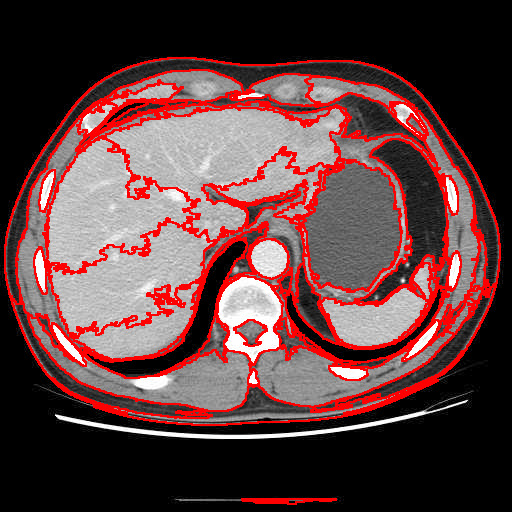} &
                     \includegraphics[width=0.18\textwidth, trim=0 60 0 50, clip]{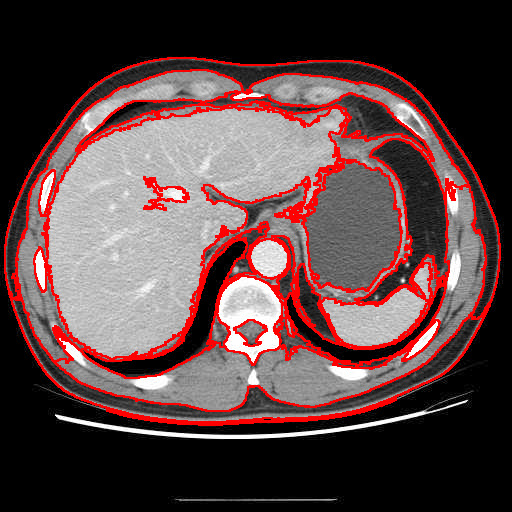} \\
                     (a) & (b) & (c) & (d) & (e)\\
                \end{tabular}
                \caption{(a) Image with the border of the liver in blue. The values of $N_0$ and $N_f$ were set to $8000$ and $20$, respectively, and all images were cropped for visualization purposes. (b) Seed preservation curve. From $8000$ to $20$ seeds, (c-d) show the DISF results when the number of superpixels are $500$ and $50$, respectively, and (e) shows its final segmentation with $20$ superpixels. } \vspace*{-0.4cm}
                \label{fig:curves}
            \end{figure*}
            
            In \cite{achanta2012slic}, the authors present a strategy (hereafter named \textit{GRID}) for selecting equally-spaced seeds in a grid pattern. For a desired number $N_f$ of superpixels, the method generates a seed set $\seedset$ whose elements are separated by $d = \sqrt{\frac{|\mathcal{\nodeset}|}{N_f}}$ from each other. For low values of $N_f$, $d$ might vary drastically with small variations in $N_f$ --- thus altering seed sampling significantly,  and consequently, superpixel segmentation. Many algorithms use this strategy for initial seed set selection~\cite{liu2018intrinsic,vargas2018isf,li2015superpixel,achanta2012slic}.
             ISF-based algorithms have exploited other types of seed sampling strategies (\eg entropy-based~\cite{vargas2018isf} and object-based~\cite{belem2019oisf, belem2019importance}), but we will focus here on state-of-the-art methods that do not take into account any prior object information, such as GRID sampling~\cite{achanta2012slic} and MIX~\cite{vargas2018isf} --- an approach which performs a GRID sampling in regions, defined by a two-level quadtree, with respect to their entropy.
             
            In order to prevent the volatility of the seed selection step and also increase the probability of a seed being in the object of interest, we propose oversampling the image (\textit{i.e.,} selecting a number $N_0 \gg N_f$ of initial seeds) with GRID since it is reasonable for a high number of seeds. In Section~\ref{subsec:seedrecomp}, we detail the necessary procedures for assuring exact $N_f$ superpixels.
            
        \subsection{IFT-based Superpixel Delineation} \label{subsec:superpcomp}
            Our algorithm generates superpixels through the Image Foresting Transform (IFT) algorithm~\cite{falcao2004ift}, which has been used for several connectivity-based operations. A path $\pi_t$ with terminus at a pixel $t\in \nodeset$ is a sequence $\langle t_1, t_2, \ldots, t_n=t\rangle$ of $n$ adjacent nodes $(t_i,t_{i+1}) \in \adjac{A}, i = 1,2,\ldots,n-1$, being \textit{trivial} when $\pi_t = \langle t \rangle$. We use $\pi_s \cdot \langle s,t \rangle$ as the extension of a path $\pi_s$ by an arc $( s,t) \in \adjac{A}$ with the two instances of $s$ being merged into one. 
            
            For a given path-cost function $f$, the IFT algorithm minimizes a \textit{cost map} $C(t) = \min_{\forall \pi_t\in \Pi} \{ f(\pi_t)\}$, by considering the set $\Pi$ of all possible paths in $G$, whenever $f$ satisfies the conditions in~\cite{ciesielski2018path} . In this process, it generates an \textit{optimum-path forest} in $G$ --- an acyclic \textit{predecessor map} $P$ that assigns to each node $t \in \nodeset$ its predecessor in the optimum path $\pi_t$ or a distinctive marker $nil \not \in \nodeset$, when $t$ is a \textit{root} of $P$. The IFT algorithm first detects the roots of the map and then finds optimum paths in a non-decreasing order of cost from the root set to the remaining nodes. As the optimum-path trees grow, they can propagate other attributes, such as the \textit{root} $R(t)$ of $t$ in $P$ and a distinct \textit{label} $L(t)$ for the tree of $t$ in $P$. In this work, we also estimate arc weights for the path-cost function based on mid-level properties of the trees during their region growing process --- an approach that was previously explored for more effective interactive object segmentation~\cite{bragantini2019dynift}.
            
            For superpixel segmentation, we define the path-cost function $f$ as follows:
            \begin{eqnarray}
            f(\langle t\rangle) & = & \left\{\begin{array}{ll}
            0 & \mbox{if $t\in \seedset \subset \nodeset$,}\\
            +\infty & \mbox{otherwise,}
            \end{array}\right. \label{eq.objcost} \nonumber \\
            f(\pi_s \cdot \langle s, t\rangle) & = & \max\{f(\pi_s),w(s,t)\},
            \end{eqnarray}
            in which $w(s,t)=\|\mu_{\tau_{R(s)}}- \textbf{I}(t)\|_2$, $\tau_x$ is a growing optimum-path tree rooted in a seed $R(s)=x\in \seedset$, and $\mu_{\tau_x}$ is the mean Lab color vector of $\tau_x$ --- \ie $\mu_{\tau_x} = \frac{1}{|\tau_x|}\sum_{\forall y\in \tau_x}  \textbf{I}(y)$. The arc weights $w(s,t)$ are dynamically estimated based on the image properties of the growing tree $\tau_{R(s)}$ at the moment $t$ is being evaluated to be part of it. In this case, $f$ satisfies the conditions in~\cite{ciesielski2018path}. However, other functions that do not satisfy those conditions have shown competitive superpixel segmentation~\cite{vargas2018isf}, given that the IFT algorithm always produces a spanning forest (the acyclic map $P$). Therefore, the seeds are meant to compete among themselves and conquer their most closely connected nodes, mathematically defining each superpixel as one optimum-path tree in $P$.
            
        \subsection{Seed Set Reduction} \label{subsec:seedrecomp}
        \begin{figure*}[t!]
                \centering
                \begin{tabular}{c}
                    \includegraphics[width=0.26\textwidth]{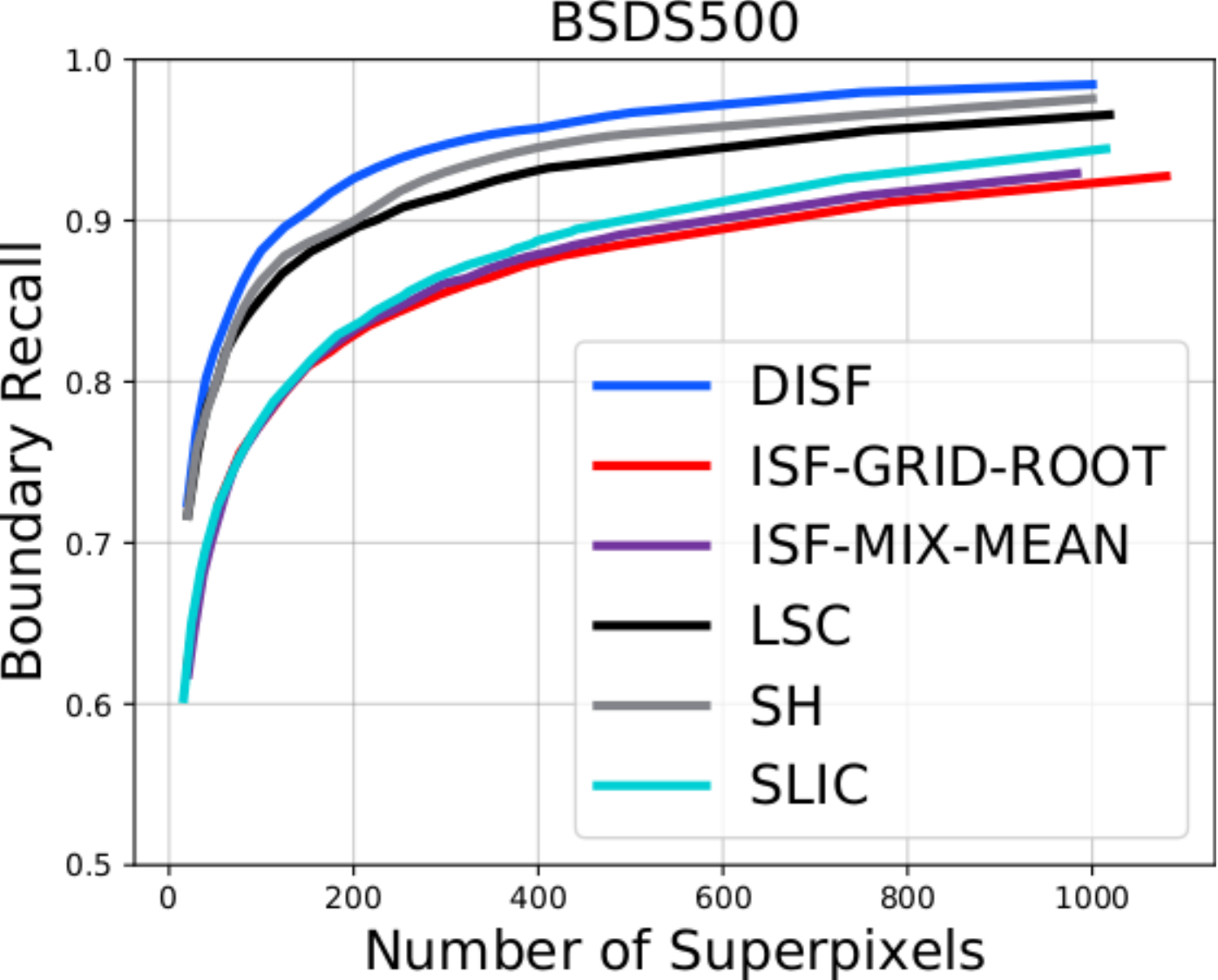}
                    \includegraphics[width=0.26\textwidth]{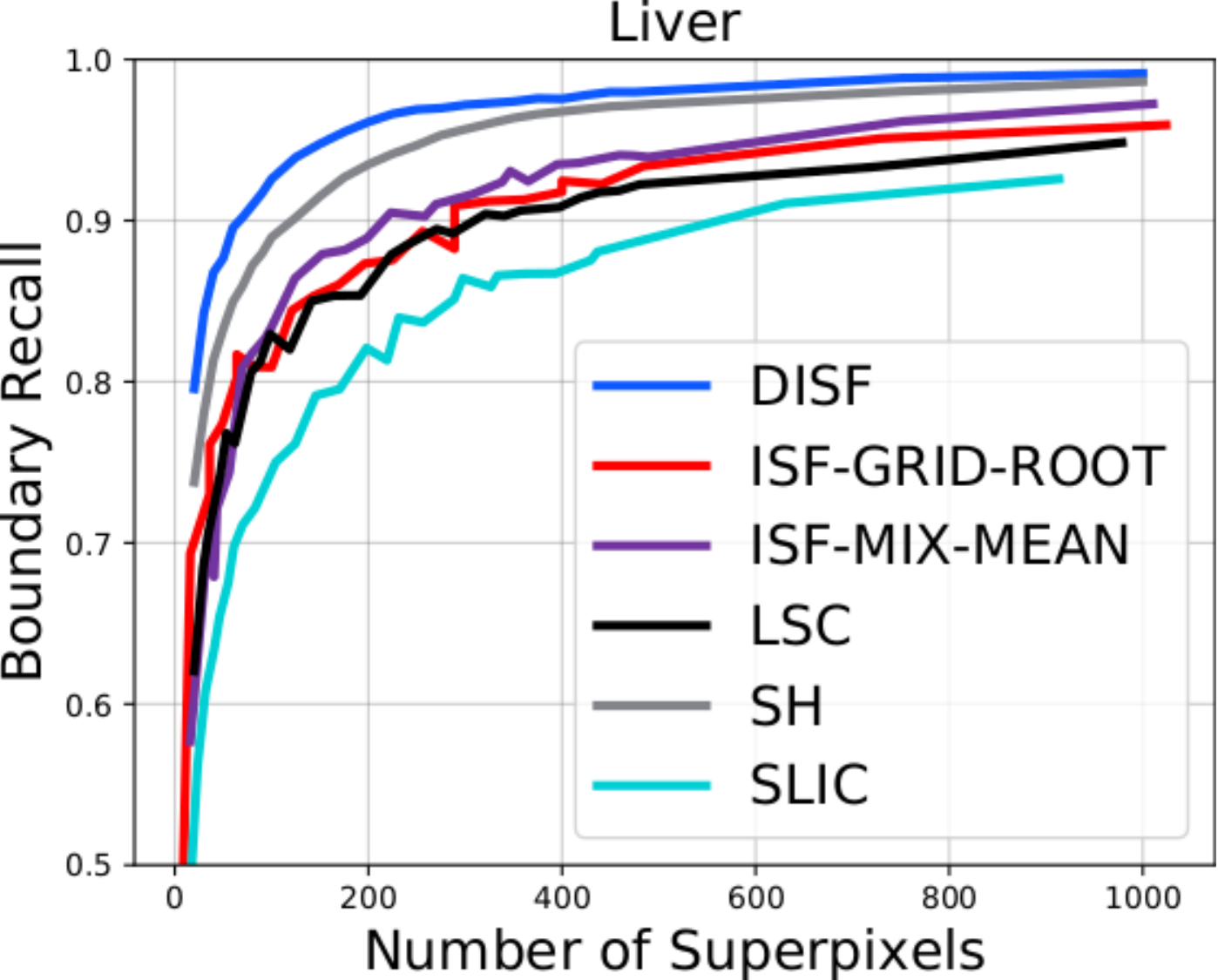}
                    \includegraphics[width=0.26\textwidth]{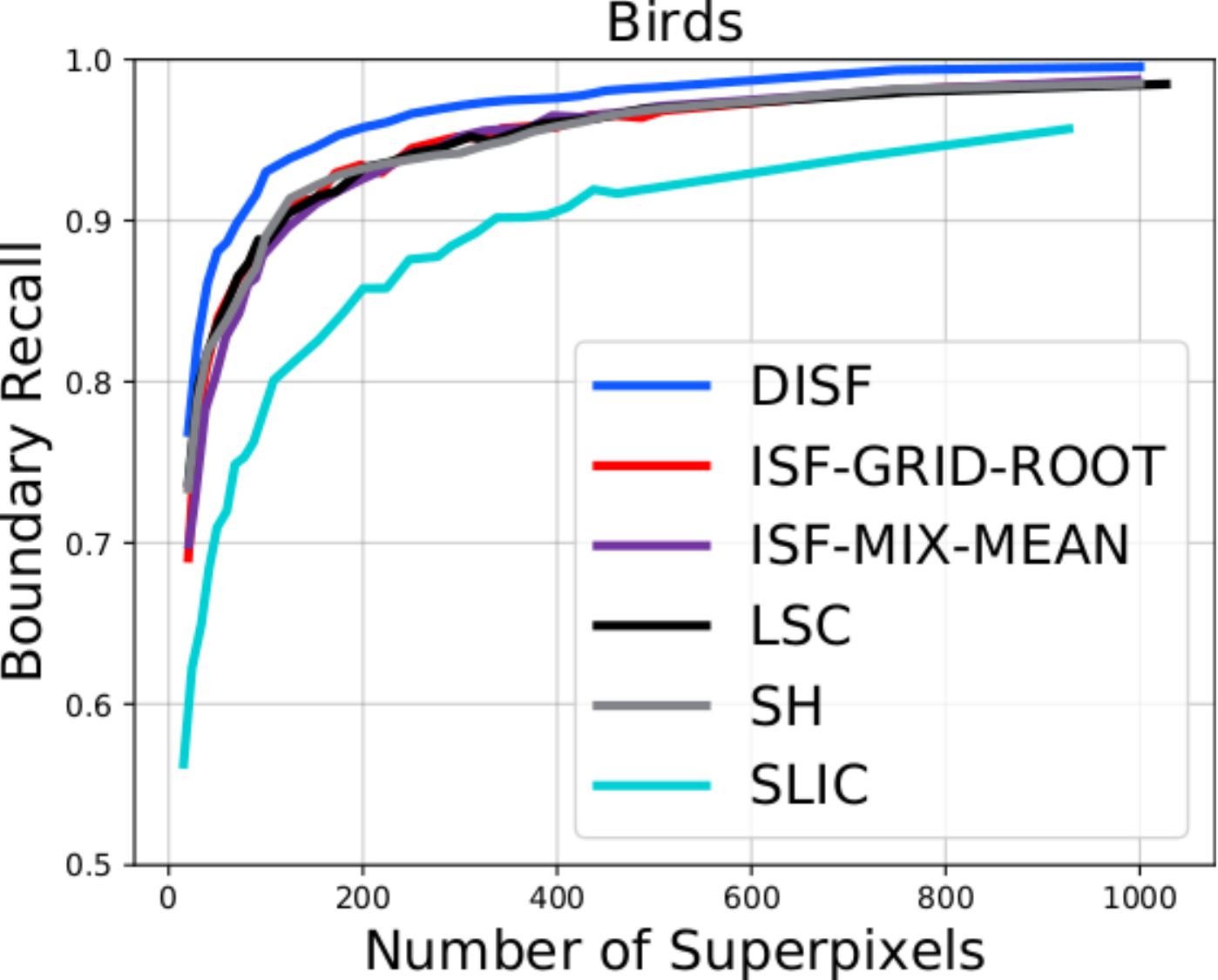} \\
                    \includegraphics[width=0.26\textwidth]{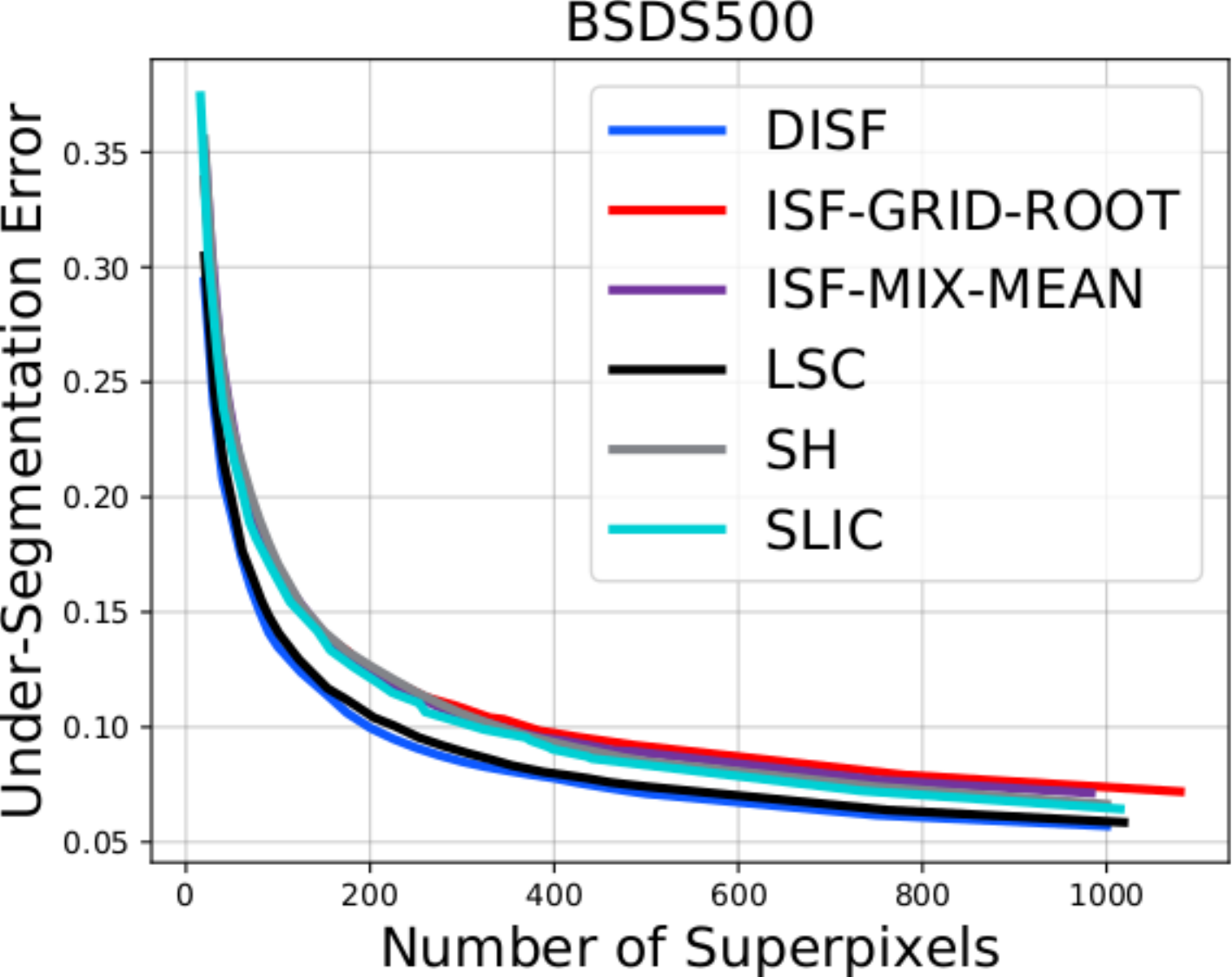} 
                    \includegraphics[width=0.26\textwidth]{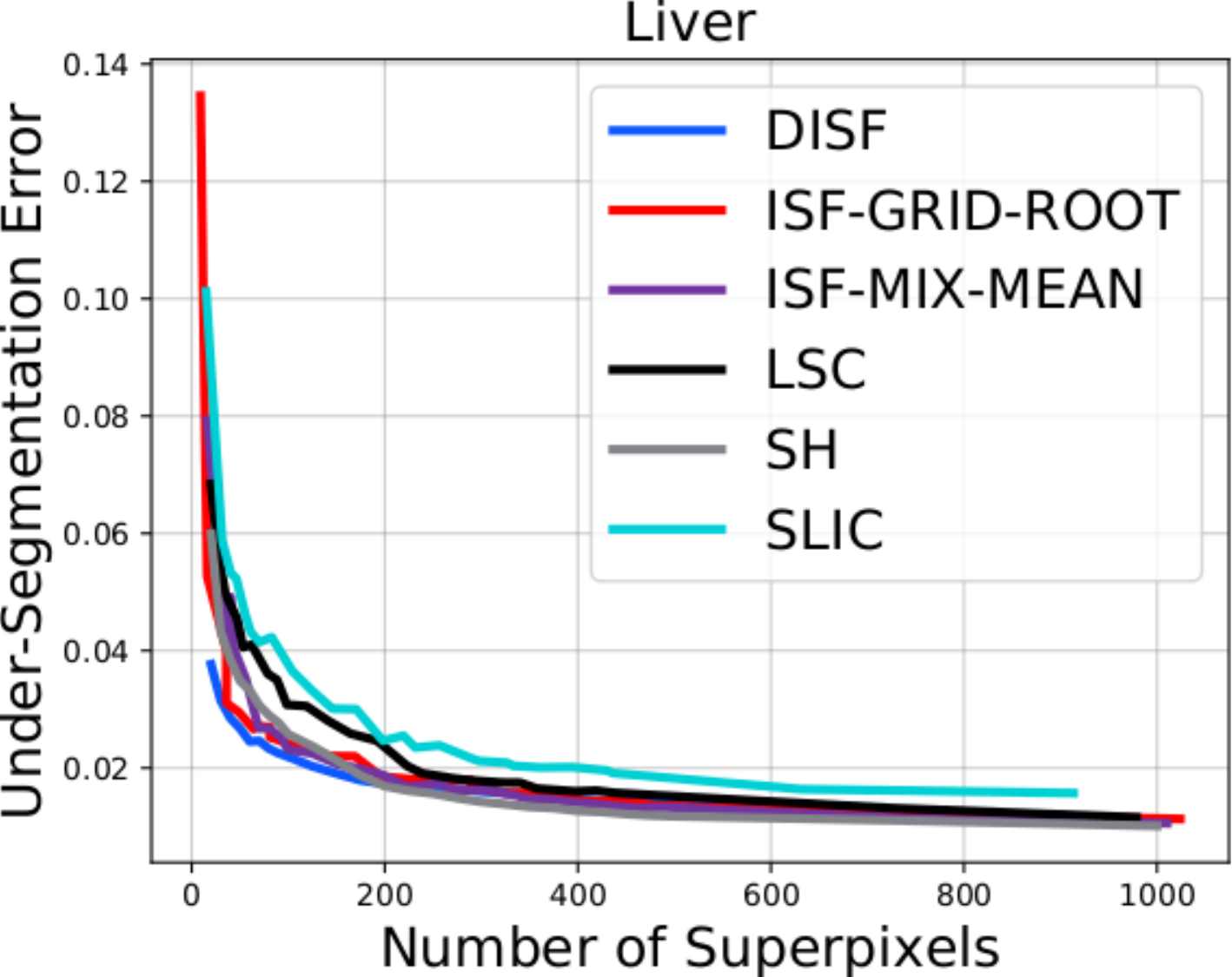} 
                    \includegraphics[width=0.26\textwidth]{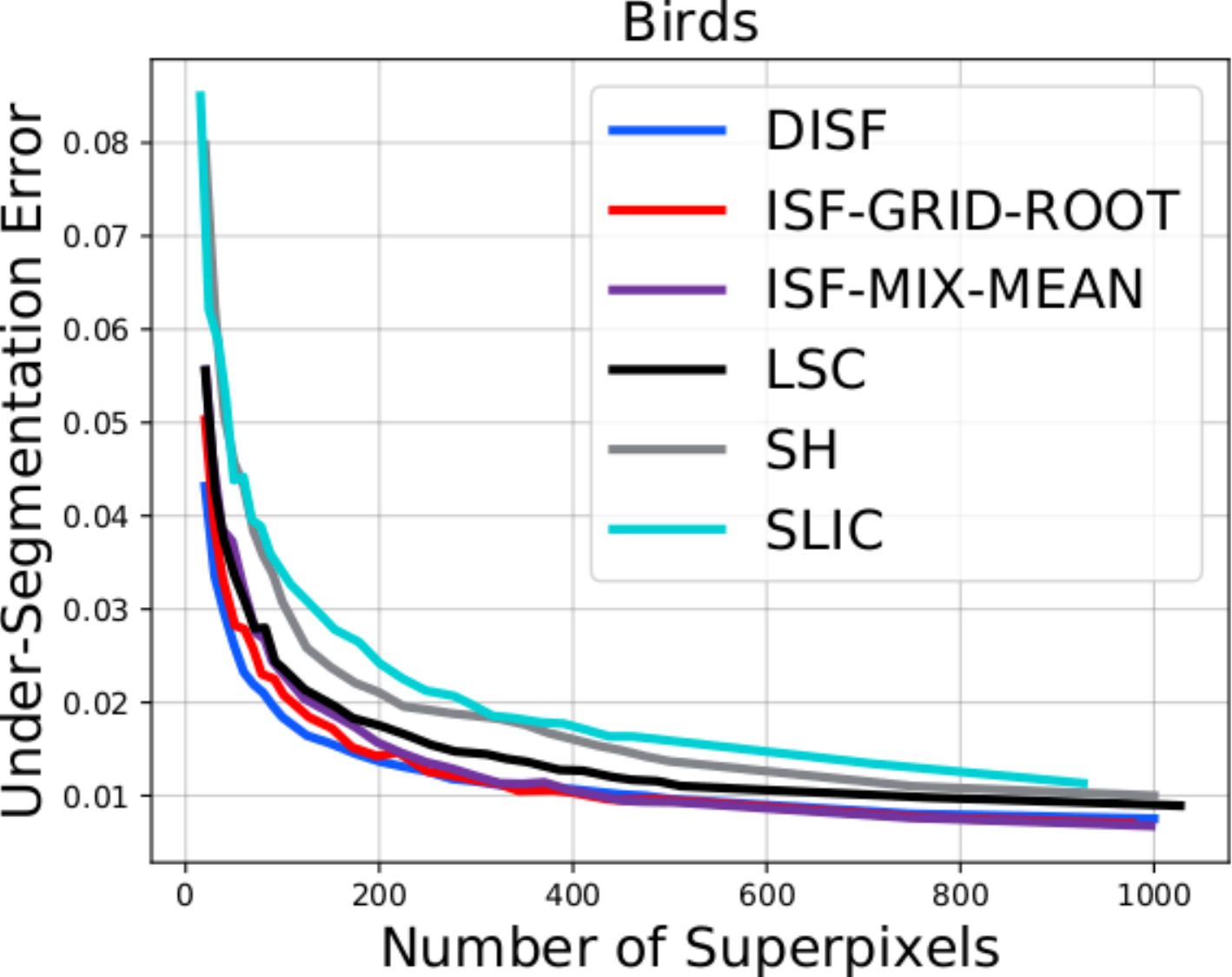}
                \end{tabular}
                \caption{Results obtained in each dataset for BR and UE. For DISF, $N_0=8000$, while for the remaining methods, the default configuration was set.}
                \label{fig:quant_anal}
            \end{figure*}
            Most recent methods start with the number of seeds equal to the number of superpixels and recompute seed location along multiple iterations by selecting the superpixel medoid --- \ie node $s \in \tau_{x} \subseteq \nodeset$ such that $\|\mu_{\tau_{x}}- \intensity(s)\|_2 \leq \|\mu_{\tau_{x}}- \intensity(t)\|_2$ for any other node $t \in \tau_x$, $x=R(s)=R(t)$ --- in order to minimize the dissimilarity between seed and remaining pixels in the superpixel (\ie maximizes homogeneity). Such strategy results in more accurate delineation, since the prevention of conquering dissimilar pixels assists in preserving important object edges. However, the size limitation of the initial seed set and such seed recomputation rule cannot assure a final set of relevant seeds to detect the important object edges. As consequence, the accuracy of those methods might be negatively affected for reduced numbers of superpixels.
            
            
            By oversampling, DISF considerably increases the chances to include all relevant seeds in the initial seed set. The challenge is to preserve those seeds in the set along the iterations. We then propose to estimate a \textit{relevance} value to each seed based on superpixel analysis.

            We intend to remove seeds that produce the smallest trees in homogeneous regions of the image by analyzing the mean color vectors of each superpixel and its neighbors. Let $\mathcal{T}$ be the set of optimum-path trees generated by the IFT algorithm. Then, a \textit{tree-adjacency relation} $\mathcal{B}$ can be defined as $\mathcal{B} = \{( \tau_x,\tau_y )\in \mathcal{T} \times \mathcal{T} \;  \; | \; \exists (s,t) \in \adjac{A}$ for some $s \in \tau_x$ and $t \in \tau_y$, with $x \neq y$ and $x,y\in \seedset$\}. The relevance $V(x)$ of a seed $x\in \mathcal{S}$ is $V(x) =   \frac{|\tau_x|}{|\mathcal{N}|}\min_{\forall (\tau_x,\tau_y) \in \mathcal{B}} \{ \|\mu_{\tau_x}- \mu_{\tau_y}\|_2\}$.
            
            Using a priority queue $Q$, every seed $x$ from the current iteration $i$, $i\in \{0,1,\ldots,T-1\}$, is inserted in $Q$ with priority $V(x)$. Then, for the next iteration $i+1$, $\seedset$ is redefined with the $M(i+1)$ seeds of highest relevance in $Q$, being the remaining ones eliminated. By that, the position of the non-removed seeds is fixed to favour the generation of similar superpixels in subsequent iterations --- \ie improving segmentation consistency by preserving the most stable object edges. 
            
            Given $N_0$ seeds at iteration $i=0$ and $N_f$ seeds as the desired number of superpixels at the last iteration $i=T-1$, the number of relevant seeds selected for iteration $i$ is defined by $M(i) = \max\{N_0\exp^{-i}, N_f\}$ (see Figure~\ref{fig:curves}). One can notice that not only such strategy does not require to provide $T$, but also takes fewer iterations for achieving effective results, as $N_f$ increases. In our experiments, for $N_0 = 8000$ and $N_f < 100$, the average number of iterations is only $5$, when other seed-based methods usually adopt $T=10$ independently of $N_f$.
            
    \section{Experimental Results} \label{sec:results}
        In this section, we present the datasets, methods, and evaluation metrics to demonstrate the results of DISF. 
    
        \subsection{Experimental Setup} \label{subsec:setup}
           
            We chose three image datasets with different object properties. \textit{Birds}~\cite{mansilla2016oriented} is a dataset with 50 natural images of birds --- colorful objects with elongated and thin parts.  \textit{Liver}~\cite{vargas2018isf} consists of 40 CT slices of the human liver --- grayscale objects with low contrast in some parts of the boundary. The test set of the popular \textit{BSDS500}~\cite{arbelaez2010contour} dataset consists of 200 natural images with a great diversity of objects.
            
            We selected five state-of-the-art superpixel segmentation algorithms for comparison, based on their object delineation performance: (i) SLIC~\cite{achanta2012slic}\footnote{https://ivrl.epfl.ch/research-2/research-current/research-superpixels/} is very popular; (ii)  SH~\cite{wei2018sh}\footnote{https://github.com/semiquark1/boruvka-superpixel} is an efficient hierarchical method; (iii) LSC~\cite{li2015superpixel}\footnote{https://jschenthu.weebly.com/projects.html} was the best for BSDS500; (iv) ISF-GRID-ROOT  --- the most competitive method in Birds~\cite{vargas2018isf}; and (v) ISF-MIX-MEAN --- the most competitive method in Liver~\cite{vargas2018isf}. For all methods, the parameter configuration recommended in the original papers was chosen.
            For evaluation, the classic metrics \textit{Boundary Recall}~(BR)~\cite{achanta2012slic} and \textit{Under-Segmentation Error}~(UE)~\cite{neubert2012superpixel} were considered, in an interval from 20 to 1000 superpixels. For any segmentation, it is desirable that superpixel borders match with object edges, and that superpixels be either inside or outside the objects. Thus, we aim higher values of BR and lower values of UE.
            
        \subsection{Comparative analysis} \label{subsec:perf_anal}

            \begin{figure}[t!]
                \centering
                \small\addtolength{\tabcolsep}{-5pt}
                \begin{tabular}{c}
                    \begin{tabular}{ccc}
                         \includegraphics[width=0.15\textwidth, trim = 170 70 100 0, clip]{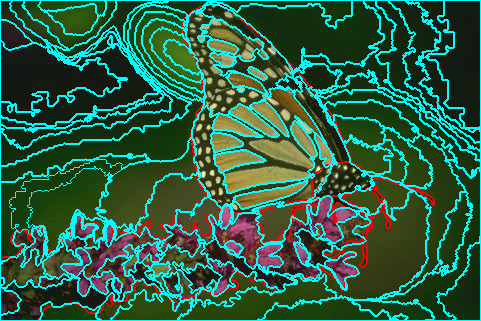} &
                        \includegraphics[width=0.15\textwidth, trim = 170 70 100 0, clip]{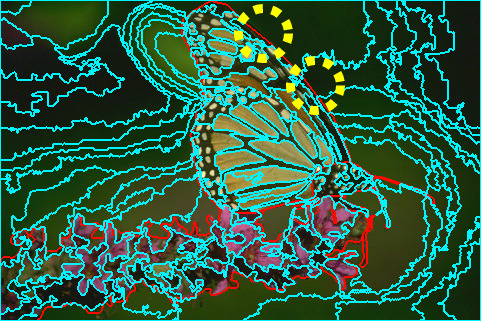} &
                        \includegraphics[width=0.15\textwidth, trim = 170 70 100 0, clip]{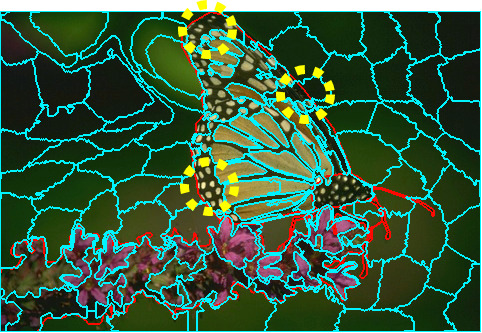}\\
                        (a) & (b) & (c)
                    \end{tabular}\\
                \end{tabular}
                \caption{Segmentation by: (a) DISF; (b) SH; and (c) LSC. The superpixel borders (in cyan) overlap the ground-truth borders (in red). For DISF, $N_0=8000$, while for the remaining methods, the default (recommended) configuration was set. The number of desired superpixels was $100$. All images were cropped for visualization purposes.} \vspace*{-0.4cm}
                \label{fig:qual_anal}
            \end{figure}
            
            In all datasets, one can see that DISF significantly outperforms all methods in BR (Figure~\ref{fig:quant_anal}), specially for $N_f < 400$. In UE, DISF is always among the best methods, specially for $N_f < 200$. Moreover, DISF presents smoother curves in all charts than the baselines, indicating  less variation in performance as $N_f$ varies --- thus, more reliable results.  
            
            When comparing DISF with its ISF-based counterparts, one may conclude that the combination among initial seed oversampling, dynamic arc-weight estimation in the path-cost function, and seed set reduction of DISF are relevant contributions for the ISF framework.  
            
            Figure~\ref{fig:qual_anal} illustrates qualitatively the superior performance of DISF over the best baselines in BSDS500. One can see that SH and LSC miss relevant object edges (in yellow). 

    \section{Conclusion and Future Work} \label{sec:concl}
        We have presented a novel three-step procedure for superpixel segmentation in the ISF framework and a new method, called DISF, that has shown to better preserve relevant object edges, specially for lower numbers of superpixels, in comparison with state-of-the-art algorithms, including one hierarchical segmentation approach based on region merging~\cite{wei2018sh}.
        
        It is worth noting that DISF cannot provide a hierarchical segmentation, but its strategy to select seeds based on relevance and its path-cost function based on dynamic arc-weight estimation can be explored in superpixel graphs for hierarchical segmentation~\cite{galvao2018risf}. Similarly, its higher effectiveness in delineation might better define symmetrical supervoxels for brain asymmetry analysis~\cite{martins2019symmisf} and class-specific superpixels for image description~\cite{castelo-fernandezLNCS19}. We also intend to investigate extensions of DISF that incorporate prior object information in the path-cost function
       ~\cite{belem2019oisf}. Thus, DISF offers several opportunities for further research and development.
    
    \bibliographystyle{abbrv}
    \bibliography{main}

\end{document}